\documentclass[10pt,twocolumn,letterpaper]{article}

\usepackage{cvpr}              

\usepackage{graphicx}
\usepackage{multirow} 
\usepackage{amsmath}
\usepackage{amssymb}
\usepackage{booktabs}
\usepackage{bbm}
\usepackage{array}
\usepackage[table,x11names]{xcolor}
\usepackage{enumitem}
\usepackage{mathrsfs}
\usepackage[accsupp]{axessibility}

\usepackage[linesnumbered,ruled,vlined,boxed]{algorithm2e}
\newcolumntype{g}{>{\columncolor{lightgray}}c}

\definecolor{lightpurple}{RGB}{150,115,166}
\definecolor{lightblue}{RGB}{108,142,191}
\usepackage[pagebackref,breaklinks,colorlinks]{hyperref}

\usepackage[capitalize]{cleveref}
\crefname{section}{Sec.}{Secs.}
\Crefname{section}{Section}{Sections}
\Crefname{table}{Table}{Tables}
\crefname{table}{Tab.}{Tabs.}

\def\cvprPaperID{7285} 
\def\confName{CVPR}
\def\confYear{2022}

\begin{document}

\title{Proto2Proto: Can you recognize the car, the way I do?}

\author{Monish Keswani \quad Sriranjani Ramakrishnan \quad Nishant Reddy \quad Vineeth N Balasubramanian\\
Indian Institute of Technology, Hyderabad\\
{\tt\small \{monish.keswani01, sriranjani.ramakrish, s.nishantreddy024\}@gmail.com, vineethnb@iith.ac.in}
}
\maketitle

\begin{abstract} 

Prototypical methods have recently gained a lot of attention due to their intrinsic interpretable nature, which is obtained through the prototypes. With growing use cases of model reuse and distillation, there is a need to also study transfer of interpretability from one model to another. We present Proto2Proto, a novel method to transfer interpretability of one prototypical part network to another via knowledge distillation. Our approach aims to add interpretability to the  ``dark" knowledge transferred from the teacher to the shallower student model. We propose two novel losses: ``Global Explanation” loss and ``Patch-Prototype Correspondence” loss to facilitate such a transfer. Global Explanation loss forces the student prototypes to be close to teacher prototypes, and Patch-Prototype Correspondence loss enforces the local representations of the student to be similar to that of the teacher. Further, we propose three novel metrics to evaluate the student’s proximity to the teacher as measures of interpretability transfer in our settings.
We qualitatively and quantitatively demonstrate the effectiveness of our method on CUB-200-2011 and Stanford Cars datasets. Our experiments show that the proposed method indeed achieves interpretability transfer from teacher to student while simultaneously exhibiting competitive performance. The code will be made available at \url{https://github.com/archmaester/proto2proto}
\end{abstract}

\section{Introduction}

\begin{figure*}
    \centering
    \includegraphics[width=1\linewidth, height=0.4\linewidth]{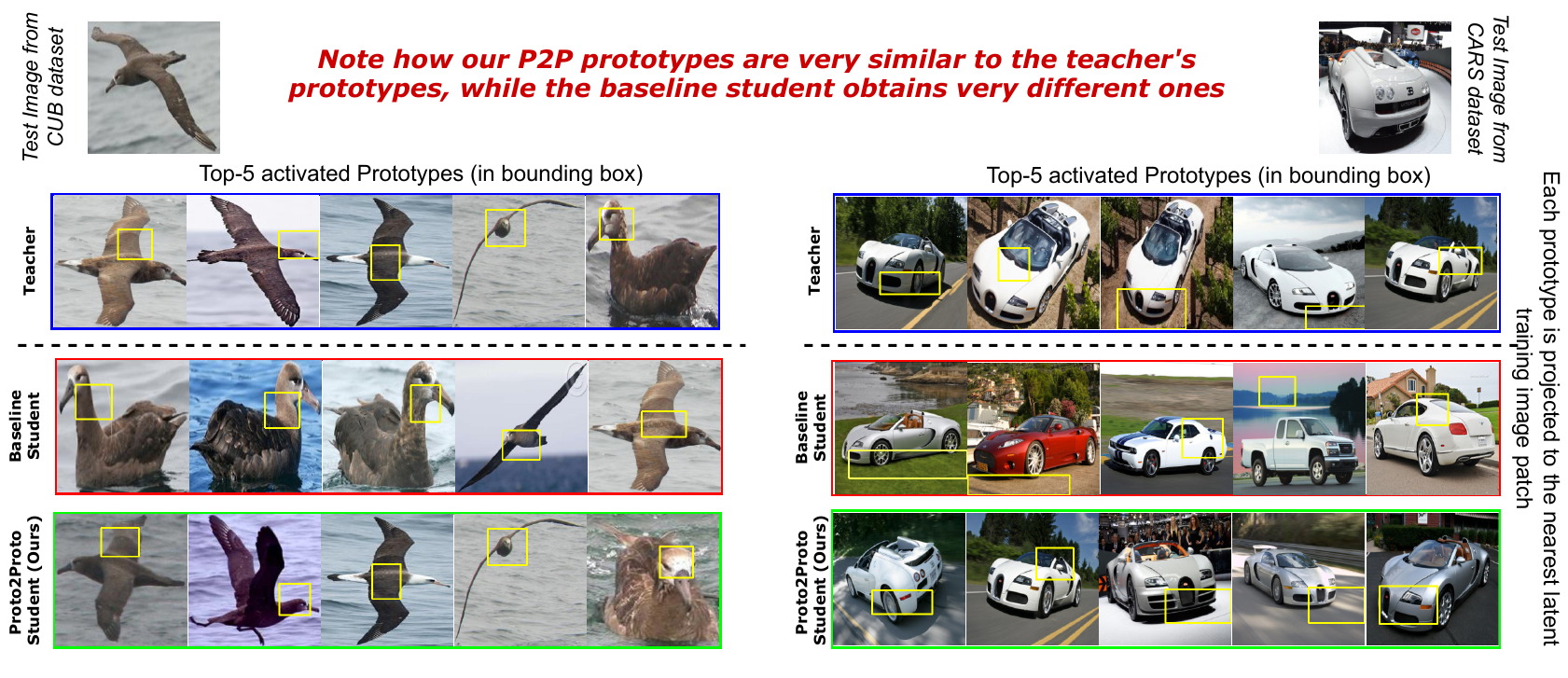}
    \caption{Comparison of sample prototypes 
    of test image between 
    Teacher, Baseline Student and Proto2Proto (P2P) Student. 
    }
    \label{fig:motivation}
\end{figure*}

Interpretability in machine learning helps humans understand the reasoning behind a model in arriving at a particular decision. From an end-user's perspective, interpretability increases the trust in the models, eventually leading to better machine learning systems, especially the ones involved in making high-stake decisions. Lipton \cite{lipton2018mythos} points out various desiderata for interpretability like trust, causality, transferability, etc. In the past decade, the primary focus of the vision community was on designing models to improve performance on tasks like classification, object detection, segmentation, etc. Deep learning models such as CNNs \cite{residual, vgg, inception} have played a considerable role in this success. However, they are often criticized as non-interpretable due to their black-box nature.

To add interpretations to the CNN models \cite{lam2021finding}, numerous efforts falling into categories of model approximation, exemplar-based, gradient-based and concept-based interpretations have been studied in recent years. In this work, we specifically focus on methods that perform model approximation using prototypes, which allow both post hoc interpretability (explaining in terms of concepts after a model is trained) and ante hoc interpretability (learning to predict and explain through prototypes jointly during training itself). 
ProtoPNet\cite{pnet} and ProtoTree\cite{ptree} are two recent methods that add interpretability to 
prototypical models. ProtoPNet learns class-specific prototypes, while ProtoTree learns class-agnostic prototypes to provide global and local interpretability. However, the efforts in this space are nascent, with no effort yet to translate to shallower networks or transfer interpretability through the learned prototypes to other models, which could have many applications in model compression, few-shot learning, continual learning, etc.

\textit{Knowledge Distillation (KD)} is a well-known technique to transfer the ``dark" knowledge from the \textit{Teacher} model to a \textit{Student} network. Many works \cite{komodakis2017paying,heo2019knowledge,ahn2019variational,tung2019similarity,peng2019correlation,tian2019contrastive,xu2020knowledge,ijcai2021-168,rkd,hinton_kd,hong2020distilling,chi2020dynamic,hong2021student,gnanasambandam2020image,Romero2015FitNetsHF} have focused on improving the performance and faithfulness of the student model to the teacher model in terms of accuracy, without much focus on the interpretability aspect. 
Liu \etal \cite{liu2018improving} distilled a black-box deep learning model to a decision tree to make it more interpretable.  Song \etal \cite{Song_2021_CVPR} constructed an intermediate decision tree to capture the intrinsic problem-solving process of the teacher and transfer it to a student. The goal of these few works was to add interpretability to a black-box teacher model through knowledge distillation. We, on the other hand, consider an already implcitly interpretable teacher model and show how our distillation method can retain faithfulness of interpretability while transferring to a student model.

To this end, we present \textbf{\textit{Proto2Proto}}, a novel method to transfer the interpretability of one prototypical part network to another. 
We consider a \textit{Teacher} network, whose interpretability we would like to transfer to a shallower \textit{Student} network. Prototype in this paper refers to a latent representation of a training image with a smaller spatial dimension called a \textit{patch}. It represents the prototypical part of images allowing finer comparisons similar to the prototypes defined in ProtoPNet \cite{pnet}. \cref{fig:motivation} illustrates the motivation of our approach. For a given test image, we visualize top-$k$ prototypes, which play the most significant role in the decision making. We compare these prototypes of teacher, baseline student and our student. As evident, our student is more faithful to the teacher in retaining similar prototypes to make decisions compared to baseline student.

In particular, we propose two novel losses: \textbf{\textit{Global Explanation loss}} and \textbf{\textit{Patch-Prototype Correspondence loss}} to achieve our objective of transferring the interpretability of the teacher to the student. In prototypical networks \cite{pnet, pshare, ptree}, knowledge is stored in the prototypes that these models learn. These prototypes can act as global explanations for the model, i.e., irrespective of the input, the model can tell which parts/regions it may focus on to make decisions. Global Explanation loss helps to transfer these global explanations or prototypes to the student. 

Similarly, for a given input, local representations obtained from the model are compared with the prototypes to determine which prototypes are present in the image. Based on the activations of prototypes, the model recognizes the image. Hence, it becomes important to generate similar activations of prototypes, for a given input, to recognize an image like the teacher. Patch-Prototype Correspondence loss helps to achieve this objective. It mimics the local representations of the teacher for which prototypes become active. Unlike \cite{Romero2015FitNetsHF}, which mimics the entire feature map of a teacher for knowledge transfer, we propose to mimic local representations of the teacher that activate prototypes.

Since this is the first such effort, to validate whether we have achieved our objective, we propose three new metrics: (i) \textbf{\textit{Average number of Active Patches (AAP)}}, which determines the average number of local representations which are active for the model. It is used to evaluate the Patch-Prototype Correspondence, with a motive of bringing this value of the student close to the teacher; (ii) \textbf{\textit{Average Jaccard Similarity of Active Patches with Teacher (AJS)}}, which determines the overlap of the active local representations of the student to the teacher. It is calculated for a pair of models, namely, teacher and student. The higher its value, the closer the student is to the teacher. It is also used to evaluate the Patch-Prototype Correspondence; and (iii) \textbf{\textit{Prototype Matching Score (PMS)}}, which evaluates how close the prototypes of the student are w.r.t. the teacher. It is used to evaluate the transfer of Global Explanations.

We summarize our contributions as follows:
\vspace{-6pt}
\begin{itemize}[leftmargin=*]
\setlength\itemsep{-0.2em}
    \item To the best of our knowledge, we present the first attempt to transfer \textit{interpretability} from a prototypical teacher to a student model. 
    \item We propose two novel losses, \textit{Global Explanation loss and Patch-Prototype Correspondence loss} for the knowledge transfer. We show that with our approach, the final layer decision module of a teacher can be used for the student directly as is, without relearning.
    \item We propose three evaluation metrics to determine the faithfulness of the student to the teacher in terms of interpretability. 
    \item We perform a comprehensive suite of experiments on benchmark datasets which show the effectiveness of our method.
\end{itemize}
\vspace{-1em}
\section{Related Works}
\subsection{Interpretability}
Interpretability for machine learning models can be provided as post-hoc explanations or as self-explanations. While the former gives intuition about the trained black-box model \cite{zeiler2014visualizing,lundberg2017unified,kim2016examples,ghorbani2019towards,kim2018interpretability,koh2017understanding}, the latter tries to understand the complex decision process by modifying architecture during training \cite{rudin2019stop,yeh2020completeness,alvarez2018towards}. Existing interpretable models can be divided into four major types \cite{lam2021finding}, namely, gradient-based, model-approximation based, conceptual interpretation and example-based methods. Our focus in this paper is on model-approximation and example-based methods.    
         Either globally or locally, the model-approximation methods approximate the representations using a self-explanatory model such as linear models and decision trees. Local models like LIME \cite{LIME}  focus on local similarity neighbourhood. While global models like soft decision trees \cite{soft_dt1,soft_dt2,soft_dt3}, adaptive neural trees \cite{adap_dt} approximate the entire deep neural models. On the other hand, example-based methods \cite{exemplar1,exemplar2} compare input images with exemplar images to interpret a single input image. Since exemplars are too specific, prototypical models \cite{prototype1,prototype2} approximate the model within a set of prototypes. These learned prototypes do not focus much on the decision process; their capacity is limited interpretability. By combining model approximation methods with prototype-based models, performance and interpretability can be handled.

Recently proposed models like ProtoPNet \cite{pnet}, ProtoTree \cite{ptree} use the above concept to improve interpretability and focus on model's decision process. ProtoPNet learns class-specific prototypes representing parts of a class approximated by a linear model for decision making. ProtoTree learns class-agnostic prototypes approximated by a decision tree making the architecture hierarchical. Our proposed work focuses on transferring the interpretable knowledge stored in these prototypes to a shallower network.

\subsection{Interpretable Knowledge Distillation}
A lot of works focus on the design and development of small models applicable in a resource-constraint deployment. Knowledge Distillation is one such model compression method to improve the performance of small student models using a large teacher model. It distils the dark knowledge by mimicking the logits \cite{caruana_kd}, or soft labels \cite{hinton_kd} from teacher to student.
Survey papers on knowledge distillation covering distillation strategies, student-teacher architectures and the recent findings can be viewed to get the background on this setup \cite{gou_kd_survey,wangkd_review}. Even though a plethora of literature for knowledge distillation is present, very few works have been proposed to focus on the interpretability aspect of knowledge distillation.

 Interpretability in knowledge distillation is commonly achieved by transferring the teacher's dark knowledge into interpretable tree-based models in one form or the other. Post-hoc interpretations were obtained for the dark knowledge by varying the inputs to trees such as matching logits \cite{impr_intel_kd}, soft targets or using different types of trees architecture viz soft decision trees \cite{soft_dt1}, vanilla decision trees , adaptive neural trees\cite{adap_dt}, Neural backed decision trees\cite{nbdt}, Gradient boosting trees\cite{gbt_intl} and generalized additive models \cite{gam_intl}. Instead of using representations from the deep neural networks, Tree-Network-Tree architecture \cite{tnt} attempts to learn a tree-based model in the input space to extract the decision path and form an embedding representation. This is further used to learn a neural network whose soft labels are used to distill knowledge into another tree-based model. This three-step process helps in making the model interpretable due to the extractable decision paths from the distilled tree. Most of the attempts above focus on relieving the tension between accuracy and interpretability. Due to the constraints on input/weight space, the knowledge could not be completely distilled into the student, or the distilled models could not be used to their full capability. 
 
Another line of thought is to use visualization methods for interpretation. DarkSight\cite{darksight} uses a simple interpretable classifier such as Naive Bayes' as a student model to mimic the dark knowledge. Applying low dimensional representation to data and jointly optimizing on model compression objective, the visualizations obtained on the network's predictions provides interpretability. 

One close work to our proposed method attempts to mimic the decision-process of the teacher in a layer-wise manner, instead of the teacher's output \cite{TDD}. They construct a decision tree using agglomerative clustering on the layers of the teacher model and use it to train the student model. Knowing the decision process in each layer makes the student model interpretable.

Our proposed work differs from the all above methods in transferring the dark knowledge (i) stored in the form of prototypes (ii) even without re-learning the decision module of the student. We also retain the faithfulness to the teacher in terms of interpretability. It also mimics the decision process made by the teacher, hence the student model can be trained to its full capacity without forgoing the performance. In terms of interpretability, the model is inherently interpretable due to the usage of an interpretable model for training a teacher. 

\section{Proto2Proto: Proposed Methodology}
\begin{figure*}[htp]
\centering
\hspace*{1.0em}
\centerline{\includegraphics[width=\linewidth, height=0.35\linewidth]{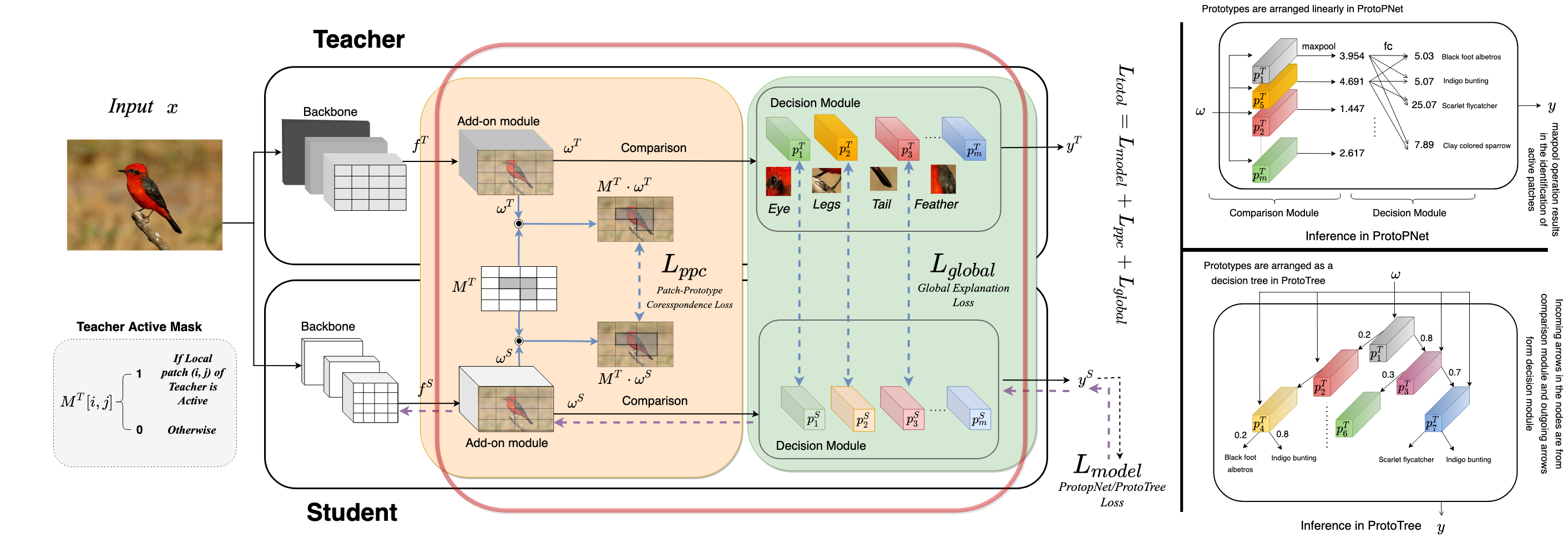}}
\caption{Our proposed architecture for training a ``Proto2Proto'' Student model via Knowledge Distillation. It shows information flow between teacher and student, as well as the inference processes {(\textbf{black} arrows: forward information flow, {\color{lightpurple}\textbf{purple}} arrows: backpropagation in student, {\color{lightblue}\textbf{blue}} arrows: loss terms introduced in our work for teacher-student alignment)}.
}
\label{fig:arch}
\end{figure*}

\begin{figure*}
\centering
\hspace*{1.0em}
\centerline{\includegraphics[width=\linewidth, height=0.15\linewidth]{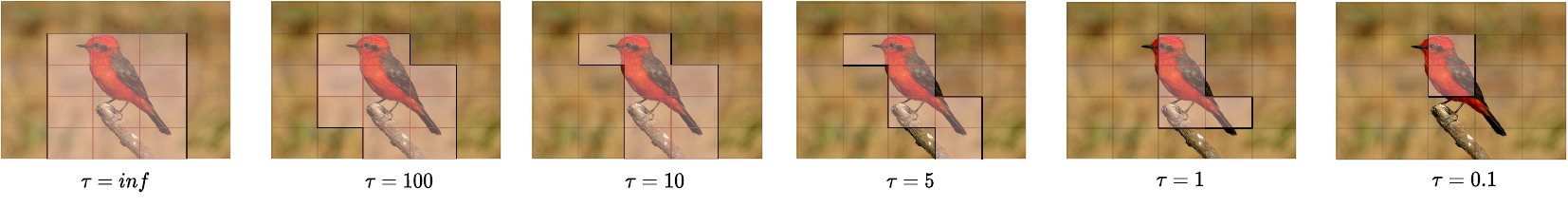}}
\caption{Effect on \textit{Active Patches} by varying $\tau$ using $L_2$ as a metric. The binary activation mask $M$ is upscaled and overlayed on the image for visualization, for different values of $\tau$. The highlighted part of the image shows the active patches.
}
\label{fig:arch_tau}
\end{figure*}

\begin{table}[htp]
\centering
\resizebox{1.0\linewidth}{!}{%
\begin{tabular}{cl}
\toprule
Notations & Description \\
\midrule
$f(x)$ & Backbone features for input $x$ \\
$\omega(x)$ & Add-on features (matched against the prototypes) for input $x$ \\
$\boldsymbol{P}$ & A tuple of prototypes \\
$p$ & Single prototype \\
$M$ & Binary mask of active patches \\
$\mathscr{L}(x)$ & Set of local patches for input $x$ \\
$l_{ij}(x)$ & ${ij}^{th}$ local patch for input $x$ \\
$\tau$ & Distance threshold to determine an active patch  \\
$C$ & Number of classes \\
\bottomrule
\end{tabular}%
}
\caption{Notations defined in the paper. The notations are superscripted by $T$ or $S$ to denote teacher or student, respectively.}
\label{tab:notation}
\vspace{-15pt}
\end{table}
Prototypical methods \cite{pnet, pshare, ptree} learn a tuple of prototypes $\boldsymbol{P}=(p_i)_{i=1}^{m}$, where $p_i \in \mathbb{R}^d$, which are representations of the most important regions in the training dataset. They facilitate a refined recognition of images. These methods usually contain four modules: backbone, add-on, comparison and decision module. The input image $x$ is first passed on to the backbone module to obtain a feature map $f(x)$ which is then passed on to an add-on module to obtain a feature map $\omega(x) \in \mathbb{R}^{H*W*d}$. The add-on module outputs $HW$ local representations or patches of dimension $d$. We denote them by a set $\mathscr{L}(x)=\{l_{11}(x), l_{12}(x),.....l_{HW}(x)\}$, where $l_{ij}(x) \in \mathbb{R}^d$. 
Note that the dimensionality of prototypes is same as local representations, i.e., $d$. 
The comparison module then compares $\mathscr{L}(x)$ with $\boldsymbol{P}$ using similarity scores. The scores between the prototypes and their nearest patch are then passed on to the decision module. In ProtoPNet \cite{pnet}, the decision module is a fully connected layer with dimensionality $\left | \boldsymbol{P}\right| * C$, where $C$ is the number of classes. Also, each prototype belongs to a particular class which makes the weights of the decision module sparse.
ProtoPShare \cite{pshare} improves on ProtoPNet and additionally merges semantically similar prototypes using novel data-dependent merge-pruning algorithm. In ProtoTree \cite{ptree}, the decision module is a decision tree where the prototypes are arranged as nodes of the decision tree. For clarity, the notations with descriptions are mentioned in \Cref{tab:notation}.\\
\vspace{-0.5pt}
    A typical information flow in prototypical models is explained as follows: a) \noindent \textbf{Inference in ProtoPNet :-} Input image $x$ is passed through CNN to obtain a set of features $\mathscr{L}(x)$. For each prototype $p$, we compute its $L_2$ distance with set $\mathscr{L}(x)$, the distances are then inverted to get similarity scores and the maximum similarity score is then computed using maxpooling. $\left|\boldsymbol{P}\right|$ similarity scores thus obtained are then fed to fully connected layer of size $\left|\boldsymbol{P}\right|*C$ to get the output logits. b) \textbf{Inference in ProtoTree :-} Here, prototypes are arranged as internal nodes of a \textit{soft, binary} decision tree (user-provided structure). Maximum similarity scores, obtained as above are then normalized to $[0,1]$ which act as probability values. For each prototype node, routing is done based on comparing max similarity score with pre-set threshold values.
Leaf nodes store the classification predictions.

\vspace{0.5em}
  In our setup, we transfer the knowledge of a trained prototypical teacher model to a student model not only to improve the performance of the student in terms of accuracy but also to bring it closer to the teacher in terms of interpretability. We denote a \textit{learned} tuple of prototypes of teacher model by $\boldsymbol{P^T}$ and a \textit{learnable} tuple of prototypes of student model by $\boldsymbol{P^S}$. 
 The global interpretability of the teacher is contained in its prototypes since they implicitly represent the parts/regions the model focuses on to make the decisions. For input image $x$, the local interpretability depends on the pair $(\mathscr{L}^{T}(x), \boldsymbol{P^T})$, since the decisions are taken based on the distance between $\boldsymbol{P^T}$ and local representations $\mathscr{L}^{T}(x)$. In order to transfer this knowledge, it is necessary that the student agrees with the teacher on local representations and prototypes. To achieve this, we 
propose two losses: \textit{Global Explanation loss} and \textit{Patch-Prototype Correspondence loss}. We first define active patches to determine which local representations 
are associated with the prototypes. Next, we define Patch-Prototype Correspondence loss, which forces the student to agree with the teacher on active local representations. Finally, we define Global Explanation loss which forces the student to agree with the teacher on the prototypes.

\vspace{-1em}
\paragraph{Active Patch:} In comparison module, each prototype is associated with the closest local patch and only the associated local patches are involved in the decision making. We call such patches as \textit{active}. For input $x$, let $k(x) \in \mathscr{L}(x)$ be a local patch. We define a function $Active$ for local patch $k(x)$ as below.

\begin{equation}
    \begin{aligned}
        \text{ If } \exists p \in \boldsymbol{P} \  \textrm{s.t. $D(k(x), p) = D^* \leq \tau$} \\
        \textrm{where $D^* = \min_{i, j} D(l_{ij}(x), p)$} \\
        \textrm{and } i,j \in \{1, 2, ....H\}, \{1, 2, ....W\}
    \end{aligned}
    \label{eq:active_argmin}
\end{equation}

\begin{equation}
    Active(k(x), \tau, \boldsymbol{P}) = 
    \begin{cases} 
          1 &  \textrm{if \cref{eq:active_argmin} is satisfied} \\
          0 & \textrm{Otherwise} \\
    \end{cases}
    \label{eq:active}
\end{equation}

\noindent where $\tau$ is a distance threshold and $\boldsymbol{P}$ is a tuple of prototypes as defined earlier. The local patch $k(x)$ is said to be active if there exists a prototype $p \in \boldsymbol{P}$ 
such that the distance of $p$ from $k(x)$ is minimum among all local patches $\mathscr{L}(x)$ and less than $\tau$. The hyperparameter $\tau$ controls the maximum distance allowed to define an active patch. \cref{fig:arch_tau} shows the effect of $\tau$ on the number of active patches. As observed, the number of active patches decrease with the value of $\tau$. We refer to $\tau$ as $\tau_{train}$ during training and $\tau_{test}$ during testing.

\vspace{-1em}
\paragraph{Patch-Prototype Correspondence Loss:} Romero \etal \cite{Romero2015FitNetsHF} proposed to mimic the local representations of the teacher as hints to the student to improve the performance of the student. 
We propose to transfer only the active local representations of the teacher's add-on layer to the student to improve its interpretability as well. 
We identify the active patches of the teacher for input $x$, and represent it as a binary activation mask $M^T$ which is defined as:
\begin{equation}
    \centering
    \begin{aligned}
    M^T(x)[i, j] = Active(l_{ij}(x), \tau_{train}, \boldsymbol{P^T}) \\
    \forall i,j \in \{1, 2, ....H\}, \{1, 2, ....W\}
    \end{aligned}
\end{equation}
We now define the Patch-Prototype Correspondence loss as:
\begin{equation}
    L_{ppc} = \frac{1}{N} \sum_{n=1}^{N} \left\Vert \  M^T(x_n) \cdot [ \omega^T(x_n) - \omega^S(x_n) ] \  \right\Vert_2
\label{eq:loss_ppc}
\end{equation}

where $N$ is the number of training images, $\omega^T(x_{n})$ and $\omega^S(x_{n})$ are the outputs of the add-on layer of teacher and student respectively, for input image $x_{n}$.

\vspace{-1em}
\paragraph{Global Explanation Loss:} To achieve our objective, the student should agree with the teacher on the prototypes. Global Explanation loss forces the prototypes of the student model to be close to that of the teacher model. We define it as:
\begin{equation}
    L_{global} = \frac{1}{m} \sum_{i=1}^{m} D(p^T_i,p^S_i)
\label{eq:loss_global}
\end{equation}
where $D$ is a distance metric (cosine, euclidean, etc.) and $m$ is the number of prototypes. We term this as  the Global Explanation loss, because the prototypes are globally interpretable, i.e, without any input we can tell which region they are focusing on. We use Euclidean distance as a metric in our experiments.

\vspace{-1em}
\paragraph{Model Loss:} It denotes the corresponding losses of the prototypical part methods \cite{pnet, ptree}. For ProtoPNet, we refer to $L_{model}$ as $L_{ppnet}$ and for ProtoTree, $L_{ptree}$.
 
The total loss is given by:
\begin{equation}
    L_{total} = L_{model} + \lambda_{global} L_{global} + \lambda_{ppc} L_{ppc}
\label{eq:loss_total}
\end{equation}
where $\lambda_{global}$ and $\lambda_{ppc}$ are the hyperparameters which are used to balance the losses.

\subsection{Evaluating Interpretability Transfer}
\label{sec:eval_metrics}
Since this is the first such effort on interpretability transfer on prototypical networks, we also propose three evaluation metrics to determine the proximity of the student model to the teacher model in terms of interpretability. Note that the aim of our evaluation metrics is not to evaluate the interpretability of individual models but to evaluate interpretability transfer between teacher and student models.

\vspace{-1em}
\paragraph{Average number of Active Patches (AAP):}
\label{sec:sec_aap}
We had earlier stated that the student should agree with the teacher on local representations and prototypes. To achieve agreement of the former, we introduced $L_{ppc}$ loss. AAP is one of the metrics which evaluates $L_{ppc}$. 
We define the average number of active patches of model $m$ as:

\begin{equation}
    \textrm{AAP}(\tau_{test}, \boldsymbol{P}) = \frac{1}{N} \sum_{n=1}^{N}\sum_{h=1}^{H} \sum_{w=1}^{W} Active(l_{hw}(x_n), \tau_{test}, \boldsymbol{P})
\end{equation}

\noindent where $N$ is the number of images, $\tau_{test}$ is the distance threshold, $\boldsymbol{P}$ is a tuple of prototypes of model $m$ (ideally, the notation should be $\boldsymbol{P^m}$, for simplicity we ignore $m$),  and $Active$ is defined in Eq. \eqref{eq:active}. The closer the value of AAP of student to that of teacher, the better the interpretability transfer.

\vspace{-1em}
\paragraph{Average Jaccard Similarity of Active Patches with Teacher (AJS):}
\label{sec:sec_ajs}
For AAP score, we counted the number of active patches. Here, we determine the overlap of the active patches of student with teacher to find out the similarity between them. AAP score is calculated for individual models whereas AJS is calculated for a (student, teacher) pair. We assign a unique identifier to all active patches of an image. For image $x$, $\mathcal{A}(x) = \{..., id_{ij}(x), ...\}$, where $id_{ij}(x) = $  \textit{UNIQUE-ID}$(ij)$ and  $Active(l_{ij}(x), \tau_{test}, \boldsymbol{P}) = 1$.
We define Average Jaccard Similarity between active patches of student $\boldsymbol{S}$ and teacher $\boldsymbol{T}$ as follows:

\begin{equation}
    \textrm{AJS}(\boldsymbol{S}, \boldsymbol{T}) = \frac{1}{N} \sum_{n=1}^{N} \frac{\left| \mathcal{A}^T(x_n) \cap \mathcal{A}^S(x_n) \right|}{\left| \mathcal{A}^T(x_n) \cup \mathcal{A}^S(x_n) \right|}
\end{equation}
where $N$ is the number of images. $\mathcal{A}^T$ and $\mathcal{A}^S$ are the active patches of teacher and student, respectively. Note that $\textrm{AJS}(\boldsymbol{T}, \boldsymbol{T})= 1$ serves as the target upper bound for the student model.

\vspace{-1em}
\paragraph{Prototype Matching Score (PMS):}
\label{sec:sec_pm}
Now, we define a metric to evaluate $L_{global}$. The intuition is to compute a matching score to measure the closeness between prototypes of teacher and student. To compute this score, the exact correspondence between teacher \& student prototypes is required, which is unknown. Hence, we use Hungarian matching algorithm (HMA) to match the prototypes. Note that HMA may not be required to match teacher prototypes with our student as there will be a sequential mapping between the two (\cref{eq:loss_global}) but is needed for baseline student.

The prototypes across models are distributed in different spaces, a direct comparison between the two using a distance metric is not trivial. Hence, we determine the similarity of prototypes across models in terms of local patches which activates the corresponding prototype.
In AJS, we maintained a list of active patches for a given input image. Here, we maintain a list of active patches across all the images, for each prototype, to determine the similarity of prototypes of two models. We use \textit{Modified-Jaccard-Similarity} as a distance metric to compare the prototypes of the teacher and the student. 
\Cref{algo_pms} summarizes the overall idea. Please refer Supplementary material for details on \textit{Modified-Jaccard-Similarity}.

\begin{algorithm}[htp]
\footnotesize{
\caption{\footnotesize{Prototype Matching Score (PMS)}}
\label{algo_pms}
\SetAlgoLined

\SetKwInOut{Input}{Input} 
\Input{$T$ - Teacher, $S$ - Student, $D$ - Test/Val Dataset}
\SetKwInOut{Output}{Output}
\Output{Prototype Matching Score (PMS)}
$numSamples = \left | D \right | $  \\
Initialize Teacher Prototype list, 
$Q^T = \left\{q^T_1, ......, q^T_m \right\}$, where $q^T_i = \phi \ \forall i \in \{1,2,...m\}, \left| \boldsymbol{P^T}\right|= m $ \\
Initialize Student Prototype list, 
$Q^S = \left\{q^S_1, ......, q^S_m \right\}$, where $q^S_i = \phi \ \forall i  \in \{1,2,...m\}, \left| \boldsymbol{P^S}\right|= m$ \\
\For{$i=1:numSamples$}{
\For{$j=1:m$}{
$k^T = \operatorname*{argmin}_{h,w} D(l^T_{hw}(x_i), p^T_j)$ \\ 
$q_j^T = q_j^T \cup \textrm{UNIQUE-ID}(k^T, x_i)$ \\ 
$k^S = \operatorname*{argmin}_{h,w} D(l^S_{hw}(x_i), p^S_j)$ \\
$q_j^S = q_j^S \cup \textrm{UNIQUE-ID}(k^S, x_i)$ \\
}
}
score-matrix $=$ \textit{Modified-Jaccard-Similarity($Q^T,Q^S$)} \\
matching-scores $=$ \textit{Hungarian}(score-matrix) \\
return \textit{Average}(matching-scores)
}
\end{algorithm}

\vspace{-2em}
\vspace{-0.25em}
\section{Experimental Results}
We demonstrate our proposed method and perform experiments on the fine-grained classification benchmark datasets CUB-200-2011 \cite{cub_birds} and Stanford Cars \cite{standford_cars}. To have fair comparison, our experimental setting is similar to the setup in ProtoPNet \cite{pnet} and ProtoTree \cite{ptree}. 
We have experimented with various architectures like ResNet \cite{residual} and VGG \cite{vgg}. For each setting, we have the results of teacher, baseline student and Proto2Proto student (Ours). The models are evaluated on interpretability using AAP, AJS and PMS as defined in \Cref{sec:eval_metrics}. Apart from this, we also evaluate all the models based on top-1 accuracy since we do not want to compromise on accuracy in favour of interpretability. 
The architectures ResNet50, and VGG19 were used for the teacher model and ResNet18, Resnet34, and VGG11 were used for the student model. Hyperparameter details can be found in the Supplementary Material.
\begin{table*}[!htbp]
    \centering
    \tabcolsep=0.3cm
    {
    \begin{tabular}{c|c|c|c c c|c}
    \toprule
    Datasets & Method & Setting & AAP & AJS ($\uparrow$) & PMS ($\uparrow$) & Top-1 Accuracy ($\uparrow$)\\ 
    \hline
    \multirow{11}{*}{\rotatebox[origin=c]{90}{CUB}}  & ProtopNet & VGG19 (Teacher) & 29.10 & 1.0 & 1.0 & 77.97 \\ 
     & ProtopNet  & VGG11 (Student) & 37.92 & 0.58 & 0.36 & 71.62 \\ 
     & Ours & VGG19 $\rightarrow$ VGG11  (KD) & 29.29 & 0.73 & 0.81 & 77.45 \\[-0.5ex]
     & & & & \textbf{(+0.15)} & \textbf{(+0.45)} & \textbf{(+5.83)} \\ 
     \cline{2-7}
     & ProtopNet & Resnet50 (Teacher)  & 20.24 & 1.0 & 1.0 & 79.22 \\ 
     & ProtopNet  & Resnet18 (Student) & 39.77 & 0.42 & 0.18 & 75.47 \\ 
     & Ours & Resnet50 $\rightarrow$ Resnet18  (KD) & 19.23 & 0.71 & 0.74 & 79.80 \\[-0.5ex]
     & & & & \textbf{(+0.29)} & \textbf{(+0.56)} & \textbf{(+4.33)} \\
    \cline{2-7}
     & ProtopNet  & Resnet50 (Teacher) & 20.24 & 1.0 & 1.0 & 79.22 \\ 
     & ProtopNet  & Resnet34 (Student) & 18.17 & 0.30 & 0.16 & 78.31 \\ 
     & Ours & Resnet50$\rightarrow$ Resnet34  (KD) & 19.33 & 0.73 & 0.79 & 79.89 \\ [-0.5ex]
     & & & & \textbf{(+0.43)} & \textbf{(+0.63)} & \textbf{(+1.6)} \\
    \hline
    \multirow{7}{*}{\rotatebox[origin=c]{90}{CARS}}    & ProtopNet & Resnet50 (Teacher) & 29.22 & 1.0 & 1.0 & 85.31 \\ 
     & ProtopNet  & Resnet18 (Student) & 32.67 & 0.45 & 0.14 & 79.96 \\ 
     & Ours & Resnet50 $\rightarrow$ Resnet18 (KD) & 29.61 & 0.62 & 0.72 & 84.00 \\ [-0.5ex]
     & & & & \textbf{(+0.17)} & \textbf{(+0.58)} & \textbf{(+4.04)} \\ 
    \cline{2-7}
     & Prototree  & Resnet50 (Teacher) & 21.35 & 1.0 & 1.0 & 85.70 \\ 
     & Prototree  & Resnet18 (Student) & 23.60 & 0.46 & 0.12 & 77.87 \\ 
     & Ours & Resnet50 $\rightarrow$ Resnet18  (KD) & 21.55 & 0.59 & 0.65 & 81.50 \\ [-0.5ex]
     & & & & \textbf{(+0.13)} & \textbf{(+0.53)} & \textbf{(+3.63)} \\ 
    \bottomrule
    \end{tabular}}
    \caption{Results of Proto2Proto student (Ours) on ProtoPNet\cite{pnet}, ProtoTree\cite{ptree} for multiple architectures like ResNet, VGG experimented on CUB and Stanford Cars. Evaluated performance using Top-1 Accuracy and interpretability using metrics AAP, AJS and PMS}
    \label{tab:table_results}
\end{table*}
    
     \Cref{tab:table_results} shows the performance of our proposed approach. Our student outperforms the baseline student in all our experiments. It is able to achieve on par or even exceed the teacher's performance in some cases. For example, VGG19 $\rightarrow$ VGG11 (KD), our method gives a 5.83\% absolute increment in accuracy compared to the baseline student. A similar trend is observed for ResNet architectures on CUB and Stanford Cars. 
     The evaluation metrics AJS will be 1 for the teacher model since the model is similar to itself. This metric shows how close the student model is to the teacher model. As seen from the results, our proposed student model is closer to the teacher in all settings compared to the baseline student. A similar argument holds for PMS, how close the student prototypes are with the teacher prototypes. The value will be 1.0 for the teacher model itself. For example, for CUB, ResNet50 $\rightarrow$ ResNet18 (KD), the increase is 0.56 in absolute values between baseline student and proposed student. The strength of our proposed student comes in the interpretability aspect along with performance. This is observed in all the experimental results as the metric AAP, AJS, and PMS are improving for our student compared to the baseline student model. For example, in ResNet50 $\rightarrow$  ResNet34 (KD), even though the performance difference is much less than other models, AJS and PMS metrics shows an increase of 0.43 and 0.63 in absolute values, respectively. Hence the proposed model is interpretable without forgoing accuracy. Additional results on the above experiments can be found in the supplementary material.
     \begin{table}[htp]
    \centering
        \scalebox{0.90}{
    \tabcolsep=0.15cm
{
    \begin{tabular}{ c c |c c c| c }
    \toprule
      Method/Setting & $\left | \boldsymbol P\right |$ & AAP & AJS ($\uparrow$) & PMS ($\uparrow$)& Acc. ($\uparrow$) \\ 
    \hline
      Baseline + PShare  & 960 & 10.69 & 0.20 & 0.13 & 70.15 \\
    \rowcolor{lightgray}
      Ours + PShare  & 960 & 13.69 & 0.33 & 0.70 & 74.01 \\ [-0.2ex]
      & & & \textbf{(+0.13)} & \textbf{(+0.57)} & \textbf{(+3.86)} \\
    \bottomrule
    \end{tabular}}}
    \caption{Results of pruning ResNet18 baseline student and our student model pruned using ProtoPShare\cite{pshare} (PShare) on Cars}
    \label{tab:table_prune}
\end{table}
\vspace{-1.0em}
\subsection{Ablation studies}
\paragraph{Ablation on losses}
 We perform experiments with various combinations of losses to demonstrate the effectiveness of our approach. \Cref{tab:table_losses} summarizes our results.
 As evident, adding the individual losses improves the performance on all metrics and all three losses combined performs even better. We additionally perform experiments by reusing the teacher's decision module for the student. It significantly boosts interpretability scores and accuracy. Notice that we get a minor benefit in interpretability scores by removing $L_{ppnet}$. However, the accuracy is more with $L_{ppnet}$. Hence, we use the last setting in our experiments.
 
 \begin{table}[htp]
    \centering
    \scalebox{0.90}{
    \tabcolsep=0.12cm
    {
    \begin{tabular}{c c c c | c c c| c}
    \toprule
     $L_{ppnet}$ & $L_{ppc}$ & $L_{global}$ & Reuse & AAP & AJS & PMS & Accuracy \\
    \hline
     \checkmark & & & & 39.77 & 0.42 & 0.18 & 75.47  \\
     \checkmark & & \checkmark & & 31.26 & 0.49 & 0.31 & 75.19  \\
     \checkmark & \checkmark & & & 30.59 & 0.54 & 0.61 & 77.63  \\
     \checkmark & \checkmark & \checkmark & & \textbf{20.89} & 0.70 & 0.69 & 78.11  \\
     \hline
    \hline
     & \checkmark & \checkmark & \checkmark & 19.29 & \textbf{0.72} & \textbf{0.76} & 79.44  \\
     \checkmark & \checkmark & \checkmark & \checkmark & 19.23 & 0.71 & 0.74 & \textbf{79.80}  \\
    \bottomrule
     \multicolumn{4}{r|}{\textit{Teacher}}  & 20.24 & 1.0 & 1.0 & 79.20  \\
     \cline{5-8}
    \end{tabular}
    }}
    \caption{Performance of ResNet18 student on different losses on CUB dataset using ResNet50 teacher . The column ''Reuse'' indicates whether we use teachers decision module for student. 
    }
    \label{tab:table_losses}
\end{table}
\vspace{-2.0em}
\paragraph{Ablation on pruning}
\label{sec:sec_prune}
ProtoPShare\cite{pshare} introduced a novel prototype-merging strategy to join semantically similar prototypes. They significantly reduced the number of prototypes of ProtoPNet model without much drop in accuracy. In \Cref{tab:table_prune}, we summarize the results of pruning. 
We prune the baseline student and Proto2Proto student from 2000 to 960 prototypes, using ProtoPshare. We observe that our Proto2Proto model significantly performs better than baseline model even after pruning.

\vspace{-1pt}
\paragraph{Comparison with KD methods}
Since our focus is on adding interpretability via knowledge distillation, to demonstrate efficacy of our method, we have compared against Relational Knowledge distillation (RKD) \cite{rkd} and Hint Loss \cite{Romero2015FitNetsHF}. Baseline student is trained with ProtoTree/ProtoPNet loss \textit{only}. Our method performs significantly well on interpretability scores as compared to the existing KD methods. In terms of accuracy, we get a minor improvement over RKD but on interpretability scores, we perform significantly better. This shows that our method performs on par with existing KD methods and enjoys interpretability for distilled dark knowledge. Additional abalation studies can be found in supplementary.
\begin{table}[htp]
    \centering
    \tabcolsep=0.15cm
    {
    \begin{tabular}{ c |c c c| c }
    \toprule
       Setting & AAP & AJS ($\uparrow$) & PMS ($\uparrow$) & Acc. ($\uparrow$) \\ 
    \hline
       Teacher & 29.22 & 1.0 & 1.0 & 85.31 \\ 
       Baseline Student & 32.67 & 0.45 & 0.14 & 79.96 \\ 
    \hline
       Hint \cite{Romero2015FitNetsHF} & 28.13 & 0.48 & 0.15 & 81.52 \\ 
       RKD \cite{rkd} & 30.85 & 0.53 & 0.27 & 83.31 \\ 
    \hline
      Ours & \textbf{29.61} & \textbf{0.62} & \textbf{0.72} & \textbf{84.00} \\ 
    \bottomrule
    \end{tabular}}
    \caption{Comparison with state-of-the-art Knowledge Distillation methods on Cars with Resnet$50$ (Teacher) and Resnet$18$ (Student)}
    \vspace{-8pt}
  \label{tab:table_kd}
\end{table}

\vspace{-12pt}

\begin{figure}
\centering
\includegraphics[width=8cm]{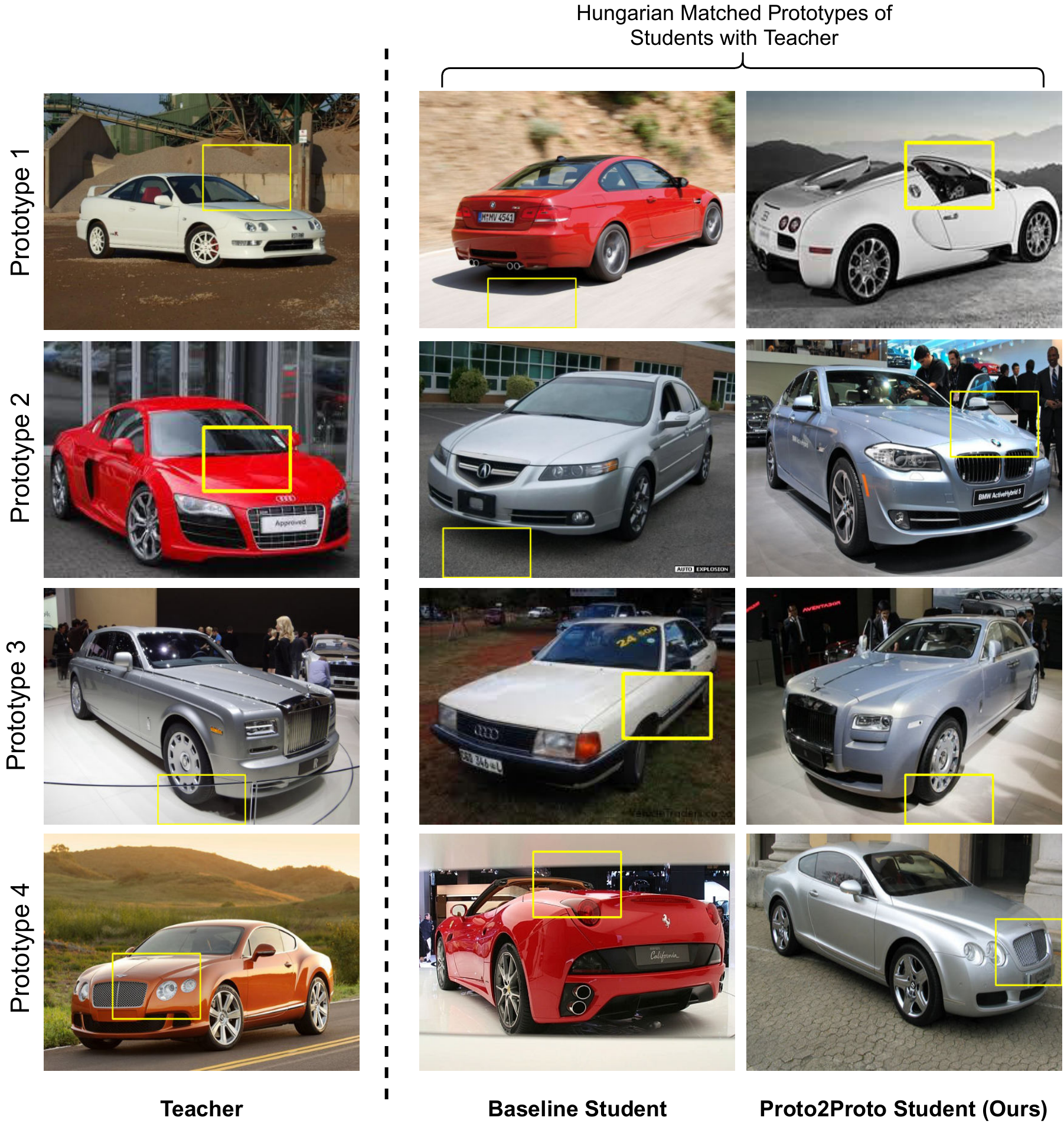}
\caption{Comparison of prototypes (each prototype projected to the nearest latent training image patch) of Teacher and Student (Baseline and Proto2Proto). The Student prototypes are matched with that of Teacher's via Hungarian Matching Algorithm.
}
\label{fig:cars_visu}
\end{figure}
\section{Visualization}
\vspace{-4pt}
\Cref{fig:cars_visu} shows the prototypes learnt by the teacher, baseline student and Proto2Proto student.
As discussed before, the prototypes of our student model are closer to the teacher model compared to the baseline student. For example, the prototype $1$ of teacher focuses on the windshield of the car and prototype $2$ focuses on the bonnet, which are also captured by the corresponding prototypes of Proto2Proto student model. The baseline student model, however, focuses on the background instead of the windshield and bonnet. As shown in  \Cref{fig:motivation}, the top-5 activated prototypes of our proposed student are similar to that of teacher's as they focuses on the wings of the bird compared to the baseline student which focuses on the neck for a given test image. This shows that our Proto2Proto model can mimic the teacher well, and the resulting prototypes retain faithfulness to the teacher compared to the baseline student. For visualizing the prototypes for ProtoPNet decision module, similar approach of using highly activated train image patch of the learnt prototype and upsampling it as described in \cite{pnet} is used.
Please refer Supplementary Material for more visualizations on ProtoTree/ProtoPNet for CUB and Cars.

\vspace{-1.0em}
\section{Conclusions and Future Work}
\vspace{-6pt}
Recent approaches have focused on designing implicit interpretable models. However, they lack the capacity to transfer knowledge in terms of interpretability. We proposed a novel framework to transfer the interpretability of teacher to student, to aid the student in making decisions like teacher. Further, we proposed three evaluation metrics to demonstrate the efficacy of our approach and reported significant performance improvement across all metrics on our student model. Extending our approach to other tasks like continual learning, transfer learning, can be an excellent future direction.

\noindent \textbf{Broader Impact and Limitations.}
Our work has no known social detrimental impact. In the current setup, we require the teacher and student to have same number of prototypes. However, the number of parameters added by the prototypes are often very less as compared to the feature extractor. Also, as shown in \Cref{sec:sec_prune}, existing pruning strategies work on our student as well which partially addresses this issue.
The proposed Prototype Matching Score uses jaccard similarity to compare the prototypes of different models. A more straightforward way would be to use a distance metric (Euclidean, cosine, etc.). However, it is non-trivial to apply such a distance metric in our setting.
Prototypical models are usually ensembled at the logits layer to obtain improved performance. Such a setup will require to evaluate interpretability between ensemble of teachers and ensemble of students. A further study is required to apply the proposed evaluation metrics in such a setup. 

\noindent \textbf{Acknowledgements.} This work has been partly supported by the funding received from DST, Govt of India, through the ICPS program as well as a Google Research Scholar Award. We thank the anonymous reviewers for their valuable feedback that improved the presentation of this paper.

{\small
\bibliographystyle{ieee_fullname}
\bibliography{paper}
}

\end{document}